\documentclass[preprint,12pt,authoryear]{elsarticle}

\usepackage{amssymb}
\usepackage{amsmath}
\usepackage{todonotes}
\usepackage{tcolorbox}
\usepackage{hyperref}
\usepackage{csquotes}

\journal{Journal of Systems and Software}

\begin{document}

\begin{frontmatter}

\title{The More the Merrier? Navigating Accuracy vs. Energy Efficiency Design Trade-Offs in Ensemble Learning Systems} 

\author[org1]{Rafiullah Omar\corref{cor1}}
\author[org2]{Justus Bogner}
\author[org1]{Henry Muccini}
\author[org2]{Patricia Lago}
\author[org3]{Silverio Martínez-Fernández}
\author[org3]{Xavier Franch}

\affiliation[org1]{
    organization = {FrAmeLab, University of L’Aquila},
    city = {L'Aquila},
    country = {Italy}
}
\affiliation[org2]{
    organization = {Vrije Universiteit Amsterdam},
    city = {Amsterdam},
    country = {The Netherlands}
}
\affiliation[org3]{
    organization = {Universitat Politècnica de Catalunya - BarcelonaTech},
    city = {Barcelona},
    country = {Spain}
}
\cortext[cor1]{Corresponding author: \url{rafiullah.omar@graduate.univaq.it}}

\begin{abstract}

\textbf{Background:}
Machine learning (ML) model composition is a popular technique to mitigate shortcomings of a single ML model and to design more effective ML-enabled systems.
Even though ensemble learning, i.e., forwarding the same request to several models and fusing their predictions, has been studied extensively regarding accuracy, we have insufficient knowledge about how to design energy-efficient ensembles.

\textbf{Objective:}
We therefore wanted to analyze three different types of design decisions for ensemble learning regarding a potential trade-off between accuracy and energy consumption:
a) ensemble size, i.e., the number of models in the ensemble, b) fusion methods (majority voting vs. a meta-model), and c) partitioning methods (whole-dataset vs. subset-based training).

\textbf{Methods:}
By combining four popular ML algorithms for classification in different ensembles, we conducted a full factorial experiment with 11 ensembles $\times$ 4 datasets $\times$ 2 fusion methods $\times$ 2 partitioning methods (176 combinations).
For each combination, we measured accuracy (F1-score) and energy consumption in J (for both training and inference).

\textbf{Results:}
While a larger ensemble size significantly increased energy consumption (size 2 ensembles consumed 37.49\% less energy than size 3 ensembles, which in turn consumed 26.96\%
less energy than the size 4 ensembles), it did not significantly increase accuracy.
Furthermore, majority voting outperformed meta-model fusion both in terms of accuracy (Cohen's $d$ of 0.38) and energy consumption (Cohen's $d$ of 0.92).
Lastly, subset-based training led to significantly lower energy consumption (Cohen's $d$ of 0.91), while training on the whole dataset did not increase accuracy significantly.

\textbf{Limitations:}
Our results cannot be easily generalized to deep learning, non-tabular ML use cases, or ensembles of substantially larger size.

\textbf{Conclusions:}
From a Green AI perspective, we recommend designing ensembles of small size (2 or maximum 3 models), using subset-based training, majority voting, and energy-efficient ML algorithms like decision trees, Naive Bayes, or KNN. Our results help to understand design decisions that impact the architectural quality concerns of environmental sustainability and functional correctness in ML-enabled systems. 
ML researchers and practitioners can use our findings to guide their design decisions for ML-enabled systems based on ensembles.

\end{abstract}


\end{frontmatter}

\section{Introduction}

ML-enabled systems, i.e., systems that use machine learning (ML) models for parts of their functionality, have experienced a rapid rise in popularity across various domains, from healthcare to finance to autonomous systems~\citep{injadat2021machine}.
However, the inherent nature of ML models poses several challenges.
These challenges include the significant computational resources required for training and inference~\citep{Schwartz2020}, potential overfitting, and the limited generalizability of model predictions for unseen data~\citep{santos2022avoiding}.
Moreover, the deployment of these models in real-world applications often leads to substantial energy consumption, raising concerns about the sustainability and cost-effectiveness of ML solutions~\citep{Strubell2019}.

To alleviate some of these issues, \textit{ML model composition} has emerged as a promising technique.
According to \citet{Apel2022}, composing multiple ML models to solve complex problems can provide a structured approach to address the lack of specifications.
For instance, instead of relying on a single monolithic model, the task of automatic image captioning can be decomposed into several steps, each handled by a different model.
This divide-and-conquer approach allows for independent development and testing of models, potentially reusing models or training data from other domains.
However, this approach is not without its own challenges, as interactions between models can lead to unexpected behaviors, necessitating a careful design to manage these interactions effectively.

From a software architecture perspective, ML model composition is an interesting consideration towards more modular ML-enabled systems.
\citet{Heyn2023} discuss the importance of adopting a compositional approach to creating architecture frameworks, particularly in distributed AI systems.
This perspective emphasizes the need for modularity and systematic design to manage the complexity of systems composed of multiple ML components.
Effective architectural frameworks can help in isolating components, designing communication channels, and implementing coordination logic to handle interactions between models.
This approach not only enhances the reliability and maintainability of ML-enabled systems but also aligns with software engineering best practices related to modularity that go back as far as to \citet{Parnas1979}.

The most common form of ML model composition is \textit{ensemble learning}.
In an ensemble, multiple ML models are trained for the same task and then their predictions on the same input are fused to improve overall accuracy~\citep{sarkar2019ensemble}.
Ensemble learning methods, such as bagging, boosting, and stacking~\citep{kunapuli2023ensemble}, aim to leverage the strengths of individual models while mitigating their weaknesses.
This approach has been widely studied for its ability to enhance the accuracy of ML predictions\citep{sagi2018ensemble, kunapuli2023ensemble}.
However, while many empirical studies have focused on the accuracy of different types of ensembles, the energy efficiency of ensemble learning has not been thoroughly explored.
Understanding a potential trade-off between accuracy and energy consumption is crucial, especially in resource-constrained environments.
Knowing how different design decisions regarding ensembles impact these quality attributes can inform the design and development of more sustainable and effective ML-enabled systems.

In this paper, we therefore present a controlled experiment to investigate how three types of design decisions influence the energy efficiency of ensembles:
a) the number of models in the ensemble, b) the fusion method, and c) the partition method for the training data.
Our goal is to support practitioners in navigating potential design trade-offs in ensemble learning.

\section{Background and Related Work}

\subsection{Energy Efficiency of ML-Enabled Systems}
The energy and carbon footprints associated with ML models and their integrated systems have become a growing concern.
For instance, training a standard natural language processing (NLP) model based on transformers can result in greenhouse gas emissions comparable to those produced by several cars over their lifetimes~\citep{Strubell2019}.
Consequently, \citet{Schwartz2020} introduced the distinction between \textit{Green AI} and \textit{Red AI}.
Traditionally, AI development focused primarily on achieving high prediction quality with little regard for energy efficiency (Red AI).
In contrast, Green AI development emphasizes minimizing energy consumption and carbon emissions while maintaining high accuracy.
While accuracy and energy consumption are often viewed as a tradeoff~\citep{Brownlee2021}, studies have demonstrated that various techniques can significantly reduce energy consumption with minimal impact on accuracy~\citep{verdecchia2022data,DelRey2023,Wei2023}.
Today, Green AI is an active research field, with a recently published review~\citep{Verdecchia2023} with data collection in 2022 identifying 98 primary studies on the topic.
Since then, Green AI research has substantially increased.
The initial development phase of ML models is typically the most energy-intensive stage within the life cycle of ML-enabled systems~\citep{Kaack2022}, thus making ML training more energy-efficient has received considerable attention.
However, other phases, such as the operation of ML-enabled systems, also offer substantial opportunities for reducing their environmental impact.
This paper focuses on the energy consumption of the training and testing phases of ensemble learning, which is an understudied area.

\subsection{Ensemble Learning}
\begin{figure}[!ht]
    \centering
    \includegraphics[width=1.0\linewidth]{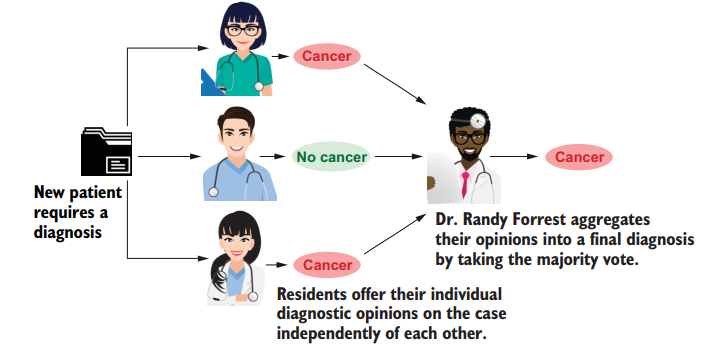} 
    \caption{Example of an ensemble of doctors, taken from \citet{kunapuli2023ensemble}}
    \label{fig:intr-doctors-ensemble}
\end{figure}

Ensemble learning, a machine learning technique that combines multiple base learners to create an ensemble learner, aims to achieve superior generalization of learning systems~\citep{dietterich2000ensemble, zhang2012ensemble}.
It is renowned for its ability to enhance model accuracy, drawing on the concept of the \enquote{Wisdom of the Crowd}~\citep{kunapuli2023ensemble, sagi2018ensemble}.
A simple analogy of ensemble learning can be illustrated by Dr. Randy Forrest's approach (see Fig.~\ref{fig:intr-doctors-ensemble}), where he consults three doctors to provide their opinions on a patient's cancer diagnosis, ultimately selecting the majority consensus as the final diagnostic~\citep{kunapuli2023ensemble}.

\begin{figure}[!ht]
    \centering
    \includegraphics[width=1.0\linewidth]{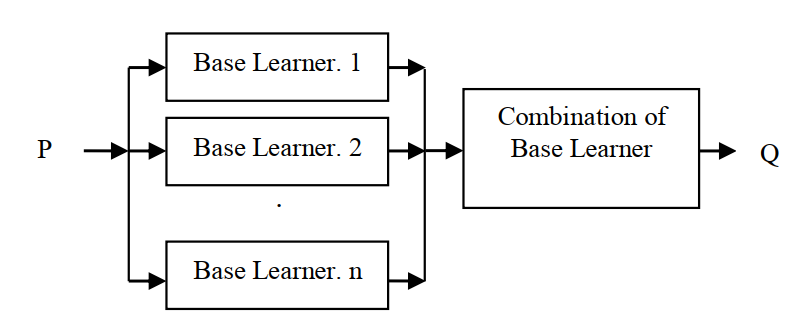} 
    \caption{Basic architecture of ensemble learning, taken from \citet{dietterich2000ensemble}}
    \label{fig:ensemble-basic-architecture}
\end{figure}

The basic architecture of ensemble learning, as shown in Fig.~\ref{fig:ensemble-basic-architecture}, involves individual models trained as \textit{base learners} (also called \textit{weak learners}).
Ensemble learning can consist of either homogeneous ensembles, constructed from a single model type, or heterogeneous ensembles, which incorporate different model types to introduce diversity~\citep{sarkar2019ensemble}.
Diverse models are typically more popular because they enhance the ensemble's robustness.
The individual models can be trained using either the entire original dataset or subsets of it.
Various methods are available for partitioning the dataset into subsets, including boosting, random forest, and cross-validation~\citep{zhang2012ensemble, wolpert1992stacked}.
We chose cross-validation to partition the dataset into subsets because it allows us to obtain representative and mutually exclusive subsets with small equal sizes.

The final component of the ensemble learning architecture involves the aggregation of the base learner outputs, which is called \textit{fusion}.
Two primary methods are employed to fuse the outputs for classification tasks: majority voting and meta-model fusion.
In majority voting, the final decision is based on the most popular option.
Potential ties are typically resolved by using the accuracy of individual models as weights~\citep{kokkinos2014breaking}.
In contrast, meta-model fusion involves training another ML model on the outputs of the base learners, with the final prediction made by the trained meta-model. Based on these different choices, we will have various combinations of the architectural components of the ensembles, leading to multiple architectural options. In our study, we will compare the accuracy and energy efficiency of these different ensemble architectures.

\subsection{Related Work}
Research on the energy consumption of software systems spans various domains, including frameworks~\citep{calero2021investigating}, data structures~\citep{oliveira2019recommending}, and programming languages~\citep{pereira2021ranking, georgiou2018your}.
This body of work, known as Green Software, aims to guide developers in creating more energy-efficient systems~\citep{anwar2020should, verdecchia2017estimating, ribeiro2021ecoandroid, cruz2019emaas}.
However, the energy efficiency of AI-based systems, particularly machine learning models, remains largely underexplored~\citep{van2021sustainable}.

\citet{Strubell2019} drew attention to the significant carbon emissions resulting from training large NLP models, prompting further investigation into AI's energy consumption.
\citet{Schwartz2020} introduced the concept of Green AI, advocating for algorithms that balance accuracy with computational costs.
\citet{bender2021dangers} emphasized the importance of curating datasets and assessing risks to mitigate the environmental impact of expanding NLP models.
Other studies have examined the impact of model size and dataset downsampling on accuracy, emphasizing AI sustainability~\citep{wu2022sustainable, zogaj2021doing}.
For instance, \citet{garcia2017identification} achieved significant energy savings with minimal accuracy loss using the Very Fast Decision Tree (VFDT) algorithm.
\citet{Georgiou2022} found that TensorFlow is more energy-efficient for training, whereas PyTorch is better for inference. \citet{omar2024sustainably} has investigated the energy efficiency of various concept drift detectors to efficiently facilitate updating ML models as needed rather than periodically\citep{jarvenpaa2024synthesis, poenaru2023retrain}.

The use of multiple ML models to improve accuracy naturally leads to a higher energy consumption in ensemble learning.
\citet{zhou2002ensembling} introduced GASEN, an approach to optimize neural network weights to select effective subsets of models.
They demonstrated that a full ensemble is not always necessary.
However, their approach itself consumes more energy because of its repetitive nature.
\citet{li2023towards} proposed the IRENE method to reduce inference costs through a learnable selector and early halting, although this method is limited to sequential processing. \citet{nijkamp2024green} introduced model selection strategies in ensemble learning to optimize performance and reduce resource consumption during the inference phase in real-time production environments.
\citet{kotary2022differentiable} presented a combinatorial optimization framework for model selection. Additionally, \citet{cordeiro2023post} proposed PS-DES, which evaluates ensembles for each query instance, thereby reducing the number of models during inference in AI pipelines.

To the best of our knowledge, none of the existing work has studied the comparative analysis of different fusion methods, the comparison of training strategies for individual models (whole dataset vs. subset), and the combination of different sizes, fusion methods, and training strategies.
Our research focuses on these aspects and analyzes the trade-offs between energy consumption and accuracy in machine learning ensembles.
By investigating the impact of ensemble size, fusion methods, and training strategies, we aim to optimize energy consumption without compromising accuracy.
This contributes to the broader goals of Green AI and sustainable machine learning practices, providing valuable insights for both researchers and practitioners.

\section{Experiment Design}
To address this gap, we designed a controlled experiment~\citep{Wohlin2012}.
In this section, we describe the most important details associated with this research method, namely our experiment objects (ML algorithms, datasets, fusion methods, and training data allocation methods), experiment variables, and experiment execution.

\subsection{Objective and Research Questions}
Driven by the need for ML-enabled systems that are both accurate and sustainable, this research investigates a potential energy consumption vs. accuracy trade-off in ensemble learning.
By analyzing the impact of ensemble size, fusion methods, and dataset partitioning, we aim to understand how these factors influence the balance between energy consumption and predictive performance.
Ultimately, accumulating evidence in this area will pave the way for guidelines that practitioners can use to navigate these trade-offs when designing ML-enabled systems.
We will achieve our research goal by answering the following research questions:

\textbf{RQ1:} How does the number of models in an ensemble influence energy consumption and accuracy?

The more models you add to an ensemble, the more energy their training and inference usage will consume.
However, adding additional models to increase diversity may also increase the overall accuracy of the ensemble\cite{kunapuli2023ensemble}, it is plausible that a trade-off exists between these two quality concerns.
We want to analyze how quickly we arrive at a point of diminishing returns, i.e., when adding another model only leads to marginal gains in accuracy while consuming substantially more energy.

\textbf{RQ2:} How do the ensemble fusion methods \textit{majority voting} and \textit{meta-model} compare in terms of energy consumption and accuracy?

We assume that using \textit{majority voting} in the fusion step does not require much energy due to the nature of the operation, but it is also fairly simplistic.
On the other hand, using a specifically trained additional ML model as a \textit{meta-model} is a more complex approach that can lead to accuracy improvements~\citep{Bonissone2011}, but may require more energy due to an additional ML model being trained.
We want to analyze how strongly this decision influences the potential overall trade-off,
especially since, with increasing ensemble size, the meta-model will be trained on more features.

\textbf{RQ3:} How do the ensemble training methods \textit{whole-dataset} and \textit{subset-based} training compare in terms of energy consumption and accuracy?

Another design decision is whether to train each individual model in the ensemble on the \textit{whole dataset} or only on a \textit{subset}.
The first will likely lead to more accuracy~\citep{prusa2015effect} but consumes more energy~\citep{salehi2023data}, while the second consumes less energy but will likely not have the same prediction performance.
We want to understand how strong the differences between these two options are, and how variation in the ensemble size moderates this relationship.

\subsection{ML Algorithms}
To increase comparability between the different experiment configurations and to limit the experiment space, we only focused on supervised classification tasks and traditional ML algorithms (no deep learning).
Various ML algorithms are available for this, each with its own strengths and weaknesses.
Combining diverse models in an ensemble may therefore yield different effects on the accuracy and energy consumption of the ensemble.
Since the choice of ML framework can also impact the energy consumption of models~\citep{Georgiou2022}, we exclusively used the widely adopted \texttt{scikit-learn} framework\footnote{\url{https://scikit-learn.org/stable/supervised_learning.html}} for comparability.
From this framework, we selected popular implemented traditional ML algorithms for classification.
Note that we excluded algorithms that are ensembles by design, e.g., random forests or the gradient boosting classifier.
The four selected ML algorithms are:

\begin{itemize}
    \item \textbf{Support Vector Machine (SVM):} SVMs are kernel-based supervised models that focus on maximizing the \enquote{margin} in classification tasks, i.e., the widest gap between the decision boundary (represented by a hyperplane) and the closest data points from each class. This margin maximization has been shown to reduce the potential for errors when the model encounters new, unseen data, i.e., it improves generalization~\citep{cristianini2000introduction}.
    
    \item \textbf{Decision Tree (DT):} DTs are supervised learning models based on tree-like structures that can be used for various classification tasks~\citep{suthaharan2016machine}. They classify instances by sorting them based on feature values. Each node in the tree represents a feature in an instance to be classified, and each branch represents a possible value for that feature. During classification, instances start at the root node and are sorted down the tree based on their features, reaching the appropriate leaf node and its corresponding class prediction~\citep{kotsiantis2007supervised}.
    
    \item \textbf{Naive Bayes (NB):} NB classifiers, known for their simplicity and efficiency, are probabilistic models based on Bayes' theorem. They operate under the assumption that each feature within a dataset contributes independently and equally to the likelihood of a sample belonging to a specific class. This means that features are mutually exclusive and have the same influence in determining the class outcome. This \enquote{naive} assumption, while not always true in reality, allows the NB classifier to be computationally fast even on large, high-dimensional datasets~\citep{misra2019machine}.
    
    \item \textbf{k-Nearest Neighbors (KNN):} the similarity-based KNN algorithm classifies new data points using the majority vote of their closest neighbors in the training data, i.e., similar data points are likely to have a similar class. It is a simple and effective method, especially for large datasets~\citep{kotsiantis2007supervised}.
\end{itemize}

We used \texttt{scikit-learn} in version 1.4.1 for all ML models in the default configuration without hyperparameter tuning.
Our primary focus is the energy consumption of ensembles, and individual model optimization via hyperparameter tuning is beyond the scope of this study.

\subsection{Datasets}
To increase the generalizability of our experiment results for classification tasks, we selected four different datasets with varying sizes and data types.
These datasets are available via the UC Irvine Machine Learning Repository, a public repository of ML training datasets that are widely used in research.

\begin{itemize}
    \item \textbf{TUNADROMD:} this IT security dataset\footnote{\url{https://archive.ics.uci.edu/dataset/813/tunadromd}} contains 4,465 software instances and 241 attributes~\citep{borah2020malware}. The target attribute for classification is a binary category that indicates whether an executable is malicious or not (malware vs. goodware).

    \item \textbf{Default of Credit Card Clients:} this finance dataset\footnote{\url{https://archive.ics.uci.edu/dataset/350/default+of+credit+card+clients}}, contains 30,000 instances and 23 attributes on default payments, demographic factors, credit data, history of payment, and bill statements of credit card clients in Taiwan from April 2005 to September 2005~\citep{misc_default_of_credit_card_clients_350}.
    
    \item \textbf{RT-IoT2022:} this Internet of Things (IoT) dataset\footnote{\url{https://archive.ics.uci.edu/dataset/942/rt-iot2022}}, derived from a real-time IoT infrastructure, contains 123,117 instances and 83 features. It simulates real-world IoT environments by combining data from various devices and mimicking cyberattacks~\citep{sharmila2023quantized}. The target attribute for the classification is a categorical variable called attack type.
    
    \item \textbf{Heart Disease:} this medical dataset\footnote{\url{https://archive.ics.uci.edu/dataset/45/heart+disease}} contains 13 features collected from 303 patients related to their cardiac features including age, gender, blood pressure, cholesterol levels, electrocardiographic (ECG) features, and more~\citep{misc_heart_disease_45}.
\end{itemize}

\subsection{Fusion Methods}
Output fusion is the step in ensemble learning where the individual predictions from the multiple base models are aggregated into one final output.
We compared two approaches within our classification scenarios:

\begin{itemize}
    \item \textbf{Majority voting:} this method is the simplest weighting method for classification problems, as the selected class is the one with the most votes~\citep{sagi2018ensemble}. For ties, we use the accuracy of individual models as weights to force a decision~\citep{gungor2021enfes}.
    
    \item \textbf{Meta-model:} in this meta-learning approach, the ensemble consists of multiple learning stages. The base models form the first stage, with their individual predictions feeding into a meta-model that generates the final output. This approach is advantageous when base models display varying effectiveness across different data subsets.
\end{itemize}

We compared these two methods in terms of energy consumption and accuracy.
For the meta-model, we chose KNN for its lower energy consumption compared to other models~\citep{verdecchia2022data}. For selecting the value of k, we followed an accepted rule of thumb that is both simple and computationally efficient: using the square root of the number of data points in the training set~\citep{zhang2017learning}.

\subsection{Dataset Partitioning Methods}
Since data-centric methods show promising potential to improve the energy efficiency of ML training~\citep{verdecchia2022data, alswaitti2024training}, we also included basic dataset partitioning into the experiment.
We selected two variations of \textit{stacking} for creating the ensemble~\citep{kunapuli2023ensemble}, i.e., each individual model was first trained using the full dataset and then using a randomly selected subset of the original dataset. 
To create these subsets, we used \textit{horizontal partitioning}~\citep{sagi2018ensemble}.
This involves dividing the dataset into disjoint subsets of equal size, where the number of subsets is equal to the number of models participating in the ensemble~\citep{ting1997stacking}. 
We chose horizontal partitioning because no single partitioning method generally outperforms others in terms of accuracy~\citep{kotsianti2007combining}.
However, reducing the size of the dataset affects the energy consumption of models during training~\citep{verdecchia2022data}.
Partitioning the dataset into horizontally disjoint subsets is a simple way of reducing the dataset size for each individual participating model.
This choice will help us understand the impact of dataset partitioning on a potential trade-off between energy consumption and accuracy in ensembles.

\subsection{Experiment Variables}
In our ensemble experiments, we analyze the trade-off between two \textbf{dependent variables}:

\begin{itemize}
    \item \textit{Ensemble energy consumption} measured in Joule (J); this includes the energy consumed for training each individual model in the ensemble (plus training the optional meta-model), but also the energy consumed during the inference to include the different fusion methods.
    \item \textit{Ensemble accuracy} measured in $F_1$ score, i.e., the harmonic mean between precision and recall
\end{itemize}

Additionally, we defined several \textbf{independent variables} that we consciously manipulated to see how they influence the dependent variables:
\begin{itemize}
    \item \textit{Ensemble size:} using our four ML algorithms, we studied ensembles of increasing size (from 1 to 4 models) for all possible combinations
    \item \textit{Fusion methods:} using majority voting vs. a meta-model
    \item \textit{Datasets:} using our four datasets that differ in number of instances and features
    \item \textit{Dataset partitioning:} using the whole dataset for training vs. horizontal partitioning
\end{itemize}

\subsection{Experiment Execution}
By combining the different independent variables, we created a \textit{full factorial design}~\citep{Wohlin2012}.
For the ensemble size, we needed to take all potential combinations of the four base models into account:

\begin{itemize}
    \setlength\itemsep{3pt}
    \item 4 individual models on their own
    \item $\binom{4}{2}=6$ ensembles with 2 models
    \item $\binom{4}{3}=4$ ensembles with 3 models
    \item 1 ensemble with 4 models
\end{itemize}

This resulted in 11 ensemble combinations.
For the individual models, we covered an experiment space of 4 ML algorithms $\times$ 4 datasets (16 combinations).
For the ensembles, we covered an experiment space of 11 ensembles $\times$ 4 datasets $\times$ 2 fusion methods $\times$ 2 partitioning methods (176 combinations).
For training the models on the entire dataset, we followed the steps outlined below for each of the 11 ensembles and 4 dataset combinations.
These four steps constitute one iteration.
We repeated each iteration 30 times to improve the reliability of the energy consumption measurements, to reduce effects of potential fluctuations in the experiment infrastructure, and to sample different parts of the dataset for training and testing for the accuracy evaluations.

\begin{enumerate}
    \item Randomly divide the dataset into three sets using a 60/20/20 split. The 60\% split set is used to train the base models, the first 20\% split set is used to test the base models, and the second 20\% split set is used to test the ensemble.
    \item Train each individual model on the training set while measuring energy consumption.
    \item Fuse the model outputs using weighting and evaluate ensemble accuracy on the testing set while measuring energy consumption.
    \item Train a meta-model and fuse the model outputs with it, then evaluate ensemble accuracy on the testing set, all while measuring energy consumption.
\end{enumerate}

In the second stage, each model was trained on a subset of the dataset using horizontal partitioning.
A similar set of steps as outlined above was followed, with the difference that the training set was randomly split into subsets of equal size according to the number of models in the ensemble.
Each model was then trained on such a randomly chosen subset instead of on the whole training set.

To instrument the experiment execution, we developed several Python scripts.
As a scripting language, Python is well suited to this task, and it is the de facto standard language for ML practitioners.
The process described above was completely automated.
To estimate the energy consumption, we used \texttt{CodeCarbon}\footnote{\url{https://codecarbon.io}}, a popular Python package.
While this is not as precise as hardware-based measurements, \texttt{CodeCarbon} underreports the actual consumed energy only by a few percentage points~\citep{Xu2023}, with a very strong correlation between both measurements ($\rho=0.94$).
Using tools like \texttt{CodeCarbon} is therefore accepted practice when the experiment goal is to compare differences in energy consumption between several variants.
The complete reproducible code is publicly available in our artifact repository.\footnote{\url{https://zenodo.org/doi/10.5281/zenodo.11664165}}
We executed the code on dedicated experiment infrastructure for software energy experiments located at the VU Amsterdam.
This infrastructure is based on a server equipped with 36 TB HDD, 384 GB RAM, and an Intel Xeon CPU including 16 cores with hyper-threading running at 2.1GHz (i.e., 32 vCPUs).
Running the experiment took approximately 50 hours.
To prevent unwanted additional load, we restricted access to the server during the experiment execution.
Before beginning the experiment, we executed a warm-up function to make the system more stable before starting data collection.
Moreover, we incorporated a 5-seconds sleep interval between each iteration to allow the infrastructure to return closer to its start state.
Together with repeating each configuration 30 times, these practices improved the reliability of our experiment and enabled a thorough and dependable assessment of the energy consumption and accuracy of the different ensembles.

\subsection{Data Analysis}
As an initial step, we assessed the normality of the distribution of results with the Shapiro-Wilk test~\citep{Shapiro1965}.
The obtained result indicated that the data exhibits non-normal distribution characteristics, as evidenced by a p-value of 9.08e-18.
To evaluate if a correlation exists between size of ensemble and energy and accuracy of ensemble, we leverage the calculation of the one-tailed Spearman’s rank correlation coefficient~\citep{Myers2002}. 
To identify significant differences in the energy consumption and accuracy among meta-model vs. majority voting fusion and subset vs. whole-set training, we employed the Mann-Whitney U test~\citep{Mann1947}, which can handle data which is not normally distributed.
To compare the magnitude of effects, we computed the percentage difference between the mean energy consumption values of different ensembles.
Furthermore, Cohen's $d$ values were computed for a more comprehensive understanding of the effect size~\citep{Cohen1988}.

\section{Results}
In this section, we present the experiment results according
to the research questions.
\subsection{Impact of Ensemble Size (RQ1)}

\begin{figure}[!ht]
    \centering
    \includegraphics[width=1.0\linewidth]{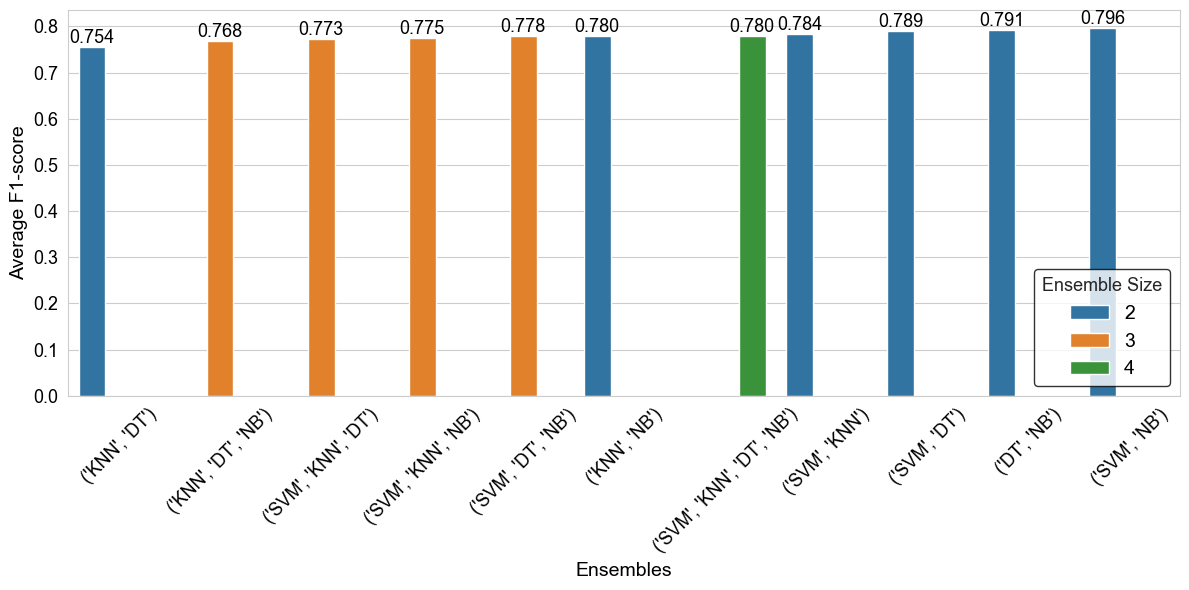} 
    \caption{Accuracy of ensembles of different sizes}
    \label{fig:rq1-accuracy-vs-size}
\end{figure}

We investigated the accuracy and energy consumption of ensembles of different sizes, specifically ensembles of sizes 2, 3, and 4.
As shown in Table~\ref{tab:rq1-size-energy-accuracy} and Fig.~\ref{fig:rq1-accuracy-vs-size}, average energy consumption clearly increases with increasing ensemble size, from 478.4 J for size 2 ensembles to 765.4 J for size 3, and up to 1,047.7 J for size 4.
The Spearman correlation test further supports these findings, yielding a significant p-value of 5.78e-35, indicating a correlation.
Specifically, an ensemble of size 2 consumes 37.49\% less energy compared to an ensemble of size 3, and an ensemble of size 3 consumes 26.96\% less energy than an ensemble of size 4.

\begin{figure}[!ht]
    \centering
    \includegraphics[width=1.0\linewidth]{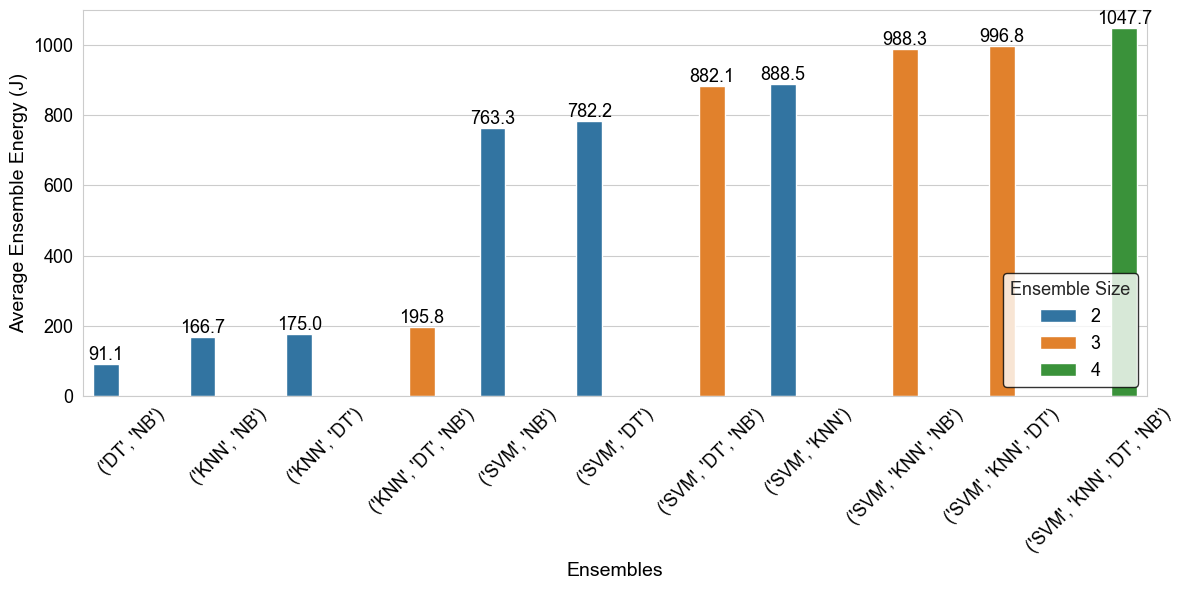} 
    \caption{Energy consumption of ensembles of different sizes}
    \label{fig:rq1-energy-vs-size}
\end{figure}

In contrast, the relationship between ensemble size and accuracy is different.
As shown in Table~\ref{tab:rq1-size-energy-accuracy} and Fig.~\ref{fig:rq1-accuracy-vs-size}, adding arbitrary models to an existing ensemble did not always increase accuracy.
In most of our studied cases, ensembles of size 2 performed better than those of sizes 3 and 4.
The Spearman correlation test yielded a p-value of 0.90 for the relationship between ensemble size and F1-score, indicating no statistically significant correlation.

This suggests that variations in the number of models within an ensemble do not meaningfully impact accuracy, as measured by the F1-score.
The accuracy difference between the worst ensemble (KNN, DT: 0.754) and the best (SVM, NB: 0.796) was only 0.042, with both having a size of 2.
The average F1-scores for ensemble sizes 2, 3, and 4 were 0.782, 0.774, and 0.780 respectively, demonstrating that increasing ensemble size does, on average, not positively impact accuracy.
However, increasing the number of models in an ensemble substantially increases energy consumption.

\begin{table}[!ht]
    \centering
    \caption{Average energy consumption vs. accuracy, Max and Min F1 are referred to Max and Min F1 of individual models in the ensemble.}
    \begin{tabular}{rrrrr}
        \textbf{Ensemble Size} & \textbf{Energy (J)} & \textbf{F1} & \textbf{Max F1} & \textbf{Min F1}  \\
        \hline
        \hline
        2 & 478.4  & 0.782 & 0.822 & 0.632  \\
        3 & 765.4 & 0.774 & 0.829 & 0.546  \\
        4 & 1047.7 & 0.780 & 0.835 & 0.474 \\
        \hline
        \hline
    \end{tabular}
    \label{tab:rq1-size-energy-accuracy}
\end{table}

\begin{tcolorbox}[colframe=black, colback=white, sharp corners]
\textbf{Answer to RQ1:} \\
On average, adding more models to an ensemble does \textbf{not} improve accuracy. 
While larger ensembles consume more energy, they are not automatically more accurate compared to smaller ones. 
Deliberate model selection to optimize energy efficiency within ensembles is therefore critical.
\end{tcolorbox}

\subsection{Impact of Majority Voting vs. Meta-Model Fusion (RQ2)}

\begin{figure*}[!ht]
    \centering
    \includegraphics[width=1.0\linewidth]{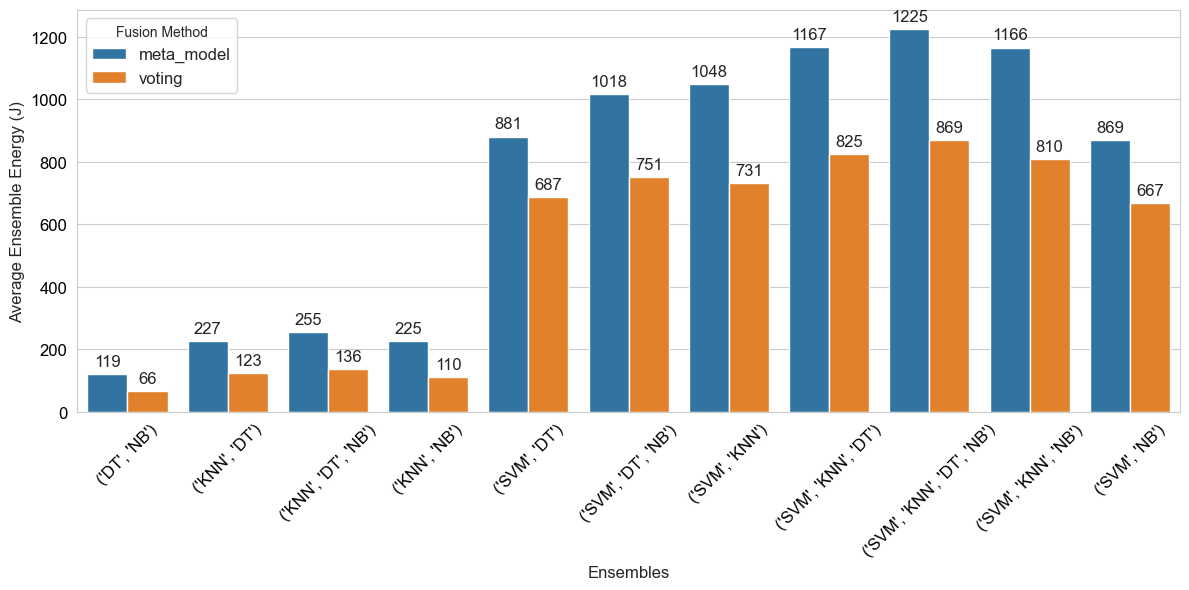} 
    \caption{Energy consumption of meta-model fusion vs. majority voting fusion}
    \label{fig:rq2-energy-size-fusion}
\end{figure*}

For RQ2, we compared the ensemble fusion methods \textit{majority voting} and \textit{meta-model} regarding their impact on energy consumption and accuracy.
As depicted in Fig.~\ref{fig:rq2-energy-size-fusion}, meta-model ensembles consumed more energy than majority voting ones in \textbf{all} cases.
The energy consumption for the meta-model method ranged from 119 J to 1,225 J, while the majority voting method ranged from 66 J to 869 J.
These findings are further supported by a Mann-Whitney U Test, which yielded a p-value of 2.66e-13 for energy consumption between these two methods, indicating a significant difference in energy consumption between the two fusion methods.

To quantify this difference, we calculated Cohen's $d$ and the percentage difference as shown in Table~\ref{tab:summary-fusion-accuracy-energy}.
The Cohen's $d$ for energy consumption was 0.92, suggesting a large effect size, with majority voting requiring significantly less energy than the meta-model approach.
The minimum individual percentage difference starts from 22.02\% in the (SVM, DT) ensemble and increases to a maximum of 51.11\% in the (KNN, DT, NB) ensemble.
Averaging the differences for ensembles of the same size, we saw a 36.16\% difference for ensembles of size 2, 33.16\% for ensembles of size 3, and 29.06\% for ensembles of size 4.
As the number of models increased in the ensembles, this difference slightly decreased.

\begin{figure*}[!ht]
    \centering
    \includegraphics[width=1.0\linewidth]{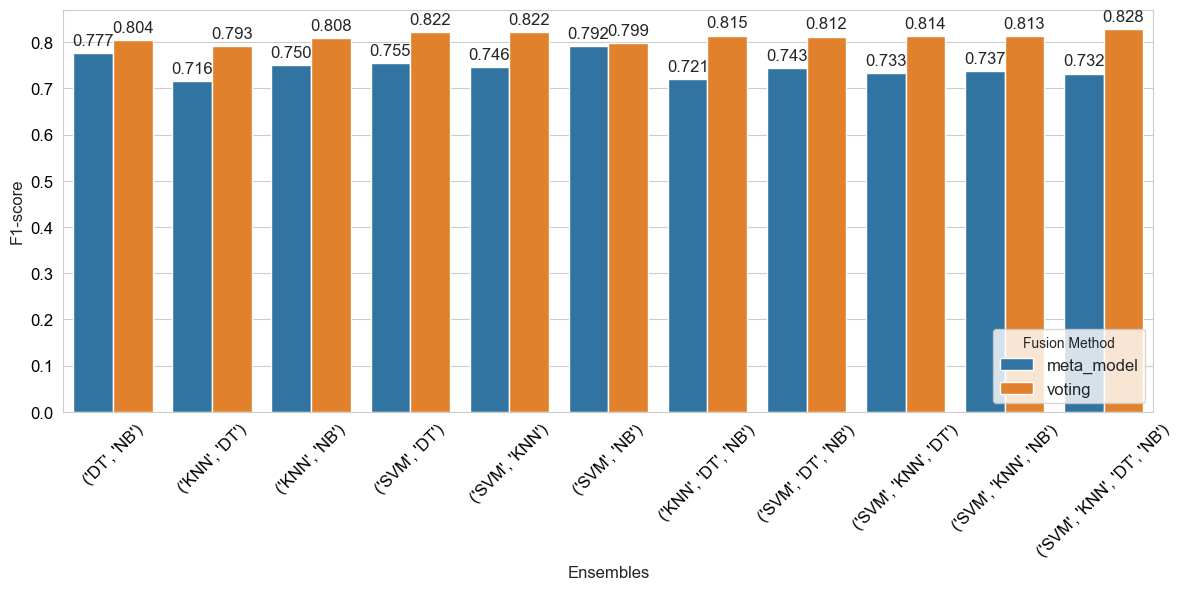} 
    \caption{Accuracy of meta-model fusion vs. majority voting fusion}
    \label{fig:rq2-accuracy-size-fusion}
\end{figure*}

Majority voting also consistently performed better than meta-model in terms of accuracy, as shown in Fig.~\ref{fig:rq2-accuracy-size-fusion}.
The best performing meta-model ensemble (SVM, NB) had an F1-score of 0.792 and the worst (KNN, DT) of 0.716, while for majority voting, the minimum was 0.793 (KNN, DT) and the maximum 0.828 (SVM, KNN, DT, NB).
These findings are further supported by a Mann-Whitney U Test, which returned a p-value of 1.33e-36, indicating a significant difference in accuracy.
To understand the effect size, we calculated Cohen's $d$ for the F1-score, which was 0.3764, indicating a small effect size.
The difference in accuracy ranged from 0.007 to 0.096.
This finding contradicts previous work~\cite{Bonissone2011}, and we will discuss in Section~\ref{sec:discussion} why this might be the case.

\begin{table*}[!ht]
    \centering
    \caption{Summary of energy consumption and accuracy of fusion methods and partition methods sorted by Fusion (J)}
    \fontsize{8}{11}\selectfont
    \begin{tabular}{lrrrrrrrrrrrr}
        ~ & \multicolumn{3}{c}{\textbf{Fusion (J)}} & 
        \multicolumn{3}{c}{\textbf{Fusion F1}} \\ \hline
        \textbf{Ensemble} & \textbf{Meta} & \textbf{M Voting} & \textbf{Diff (\%)} & \textbf{Meta} & \textbf{Majority voting} & \textbf{Diff} \\ \hline
        \hline
        {\hspace{-0.2cm}}('KNN', 'NB') & 225 & 110 & 51.11 & 0.750 & 0.808 & 0.058  \\ 
        {\hspace{-0.2cm}}('KNN', 'DT', 'NB') & 255 & 136 & 46.67 & 0.721 & 0.815 & 0.094 \\ 
        {\hspace{-0.2cm}}('KNN', 'DT') & 227 & 123 & 45.81 & 0.716 & 0.793 & 0.077  \\ 
        {\hspace{-0.2cm}}('DT', 'NB') & 119 & 66 & 44.54 & 0.777 & 0.804 & 0.027 \\
        {\hspace{-0.2cm}}('SVM', 'KNN', 'NB') & 1,166 & 810 & 30.53 & 0.737 & 0.813 & 0.076 \\ 
        {\hspace{-0.2cm}}('SVM', 'KNN') & 1,048 & 731 & 30.25 & 0.746 & 0.822 & 0.076 \\
        {\hspace{-0.2cm}}('SVM', 'KNN', 'DT') & 1,167 & 825 & 29.31 & 0.733 & 0.814 & 0.081  \\
        {\hspace{-0.2cm}}('SVM', 'KNN', 'DT', 'NB') & 1,225 & 869 & 29.06 & 0.732 & 0.828 & 0.096  \\ 
        {\hspace{-0.2cm}}('SVM', 'DT', 'NB') & 1,018 & 751 & 26.23 & 0.743 & 0.812 & 0.069  \\ 
        {\hspace{-0.2cm}}('SVM', ' NB') & 869 & 667 & 23.25 & 0.792 & 0.799 & 0.007 6 \\
        {\hspace{-0.2cm}}('SVM', 'DT') & 881 & 687 & 22.02 & 0.755 & 0.822 & 0.067  \\ 
        \hline
        \hline
    \end{tabular}
    \label{tab:summary-fusion-accuracy-energy}
\end{table*}
\begin{tcolorbox}[colframe=black, colback=white, sharp corners]
\textbf{Answer to RQ2:} \\
Majority voting emerged as the more energy-efficient fusion method in comparison to meta-model.
It not only required on average 34.43\% less energy but also increased accuracy by 0.066.
In our experiment, majority voting was always preferable to meta-model fusion.
\end{tcolorbox}

\subsection{Impact of Whole-Dataset vs. Subset-Based Training (RQ3)}
\begin{figure*}[!ht]
    \centering
    \includegraphics[width=1.0\linewidth]{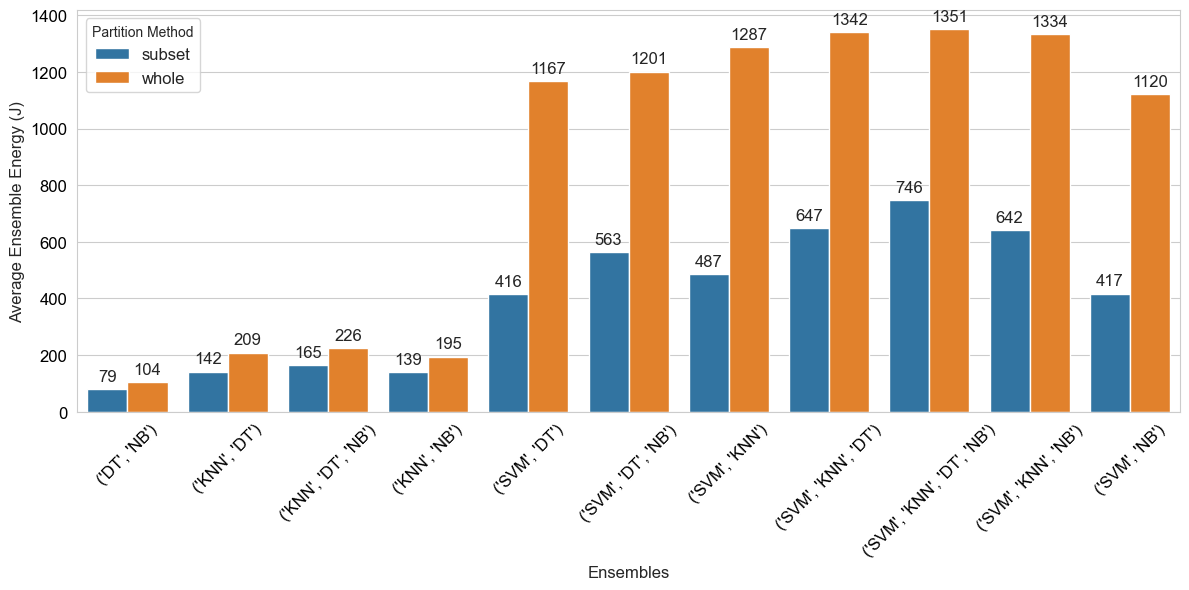} 
    \caption{Energy consumption of whole-dataset vs. subset-based training}
    \label{fig:rq3-energy-partition}
\end{figure*}

Regarding training set partition methods, Fig.~\ref{fig:rq3-energy-partition} reveals clear differences.
Training the ensemble models on the whole dataset consistently consumed more energy than training on subsets.
This was further supported by a Mann-Whitney U test with a p-value of 2.3e-17, which indicates a significant difference.
To understand the effect size, we calculated Cohen's $d$ and obtained 0.91, indicating a large effect.
Additionally, we calculated the percentage difference: for individual ensembles, the difference starts from 24.04\% and increases to 64.35\%.
On average, in ensembles of size 2, training the models on subsets saved 45.68\% energy; in ensembles of size 3, subsets saved 45.94\%; and in ensembles of size 4, training on subsets saved 44.78\% energy.
For all ensemble sizes, training on subsets saved on average 45.7\% energy.

\begin{table*}[!ht]
    \centering
    \caption{Summary of energy consumption and accuracy of fusion methods and partition methods sorted by Partition (J)}
    \fontsize{8}{11}\selectfont
    \begin{tabular}{lrrrrrrrrrrrr}
        ~ &  \multicolumn{3}{c}{\textbf{Partition (J)}} & \multicolumn{3}{c}{\textbf{Partition F1}} \\ \hline
        \textbf{Ensemble} & \textbf{Subset} & \textbf{Whole} & \textbf{Diff (\%)} & \textbf{Subset} & \textbf{Whole} & \textbf{Diff} \\ \hline
        \hline
        {\hspace{-0.2cm}}('SVM', 'DT') & 416 & 1,167 & 64.35 & 0.779 & 0.801 & 0.022 \\ 
        {\hspace{-0.2cm}}('SVM', ' NB') & 417 & 1,120 & 62.77 & 0.788 & 0.804 & 0.016 \\
        {\hspace{-0.2cm}}('SVM', 'KNN') & 487 & 1,287 & 62.16 & 0.778 & 0.791 & 0.013 \\
        {\hspace{-0.2cm}}('SVM', 'DT', 'NB') & 563 & 1,201 & 53.12 & 0.776 & 0.781 & 0.005 \\ 
        {\hspace{-0.2cm}}('SVM', 'KNN', 'NB') & 642 & 1,334 & 51.87 & 0.775 & 0.775 & 0.000 \\ 
        {\hspace{-0.2cm}}('SVM', 'KNN', 'DT') & 647 & 1,342 & 51.79 & 0.773 & 0.774 & 0.001 \\ 
        {\hspace{-0.2cm}}('SVM', 'KNN', 'DT', 'NB') & 746 & 1,351 & 44.78 & 0.783 & 0.778 & -0.005 \\ 
        {\hspace{-0.2cm}}('KNN', 'DT') &  142 & 209 & 32.06 & 0.747 & 0.762 & 0.015 \\ 
        {\hspace{-0.2cm}}('KNN', 'NB') & 139 & 195 & 28.72 & 0.772 & 0.788 & 0.016 \\ 
        {\hspace{-0.2cm}}('KNN', 'DT', 'NB') & 165 & 226 & 26.99 & 0.767 & 0.768 & 0.001 \\         
        {\hspace{-0.2cm}}('DT', 'NB') & 79 & 104 & 24.04 & 0.781 & 0.802 & 0.021 \\

        \hline
        \hline
    \end{tabular}
    \label{tab:summary-partition-accuracy-energy}
\end{table*}

\begin{figure*}[!ht]
    \centering
    \includegraphics[width=1.0\linewidth]{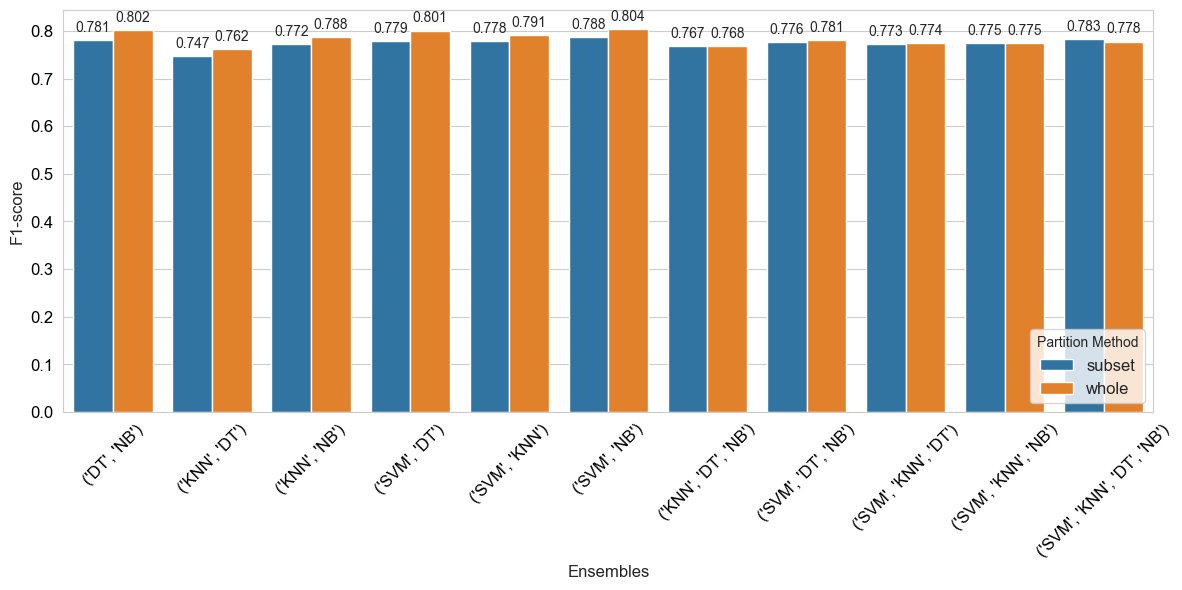} 
    \caption{Accuracy of whole-dataset vs. subset-based training}
    \label{fig:rq3-accuracy-partition}
\end{figure*}

When comparing the F1-scores, Fig.~\ref{fig:rq3-accuracy-partition} shows that training on the whole dataset either gave the same or higher accuracy compared to subsets.
The only exception was for ensembles of size 4, where subsets offered an average increase of 0.005 in accuracy compared to the whole dataset.
Even though this difference in accuracy is very small, the Mann-Whitney U test still led to a significant p-value of 0.2e-3 due to the large sample size.
However, the Cohen's $d$ value of 0.02, shows that the effect size is close to negligible.

To quantify the difference, we calculated the differences as shown in Table~\ref{tab:summary-partition-accuracy-energy}, and found that the difference in accuracy starts from 0 and increases to 0.022.
On average, training on the whole dataset offered a 0.0095 increase in accuracy compared to training the model on subsets.

\begin{tcolorbox}[colframe=black, colback=white, sharp corners]
\textbf{Answer to RQ3:} \\
Whole-dataset training consumed on average 45.7\% more energy, but offered only a negligible (0.0095) increase in accuracy compared to subset-based training.
\end{tcolorbox}

\section{Discussion}
\label{sec:discussion}
\begin{figure}[!ht]
    \centering
    \includegraphics[width=1.0\linewidth]{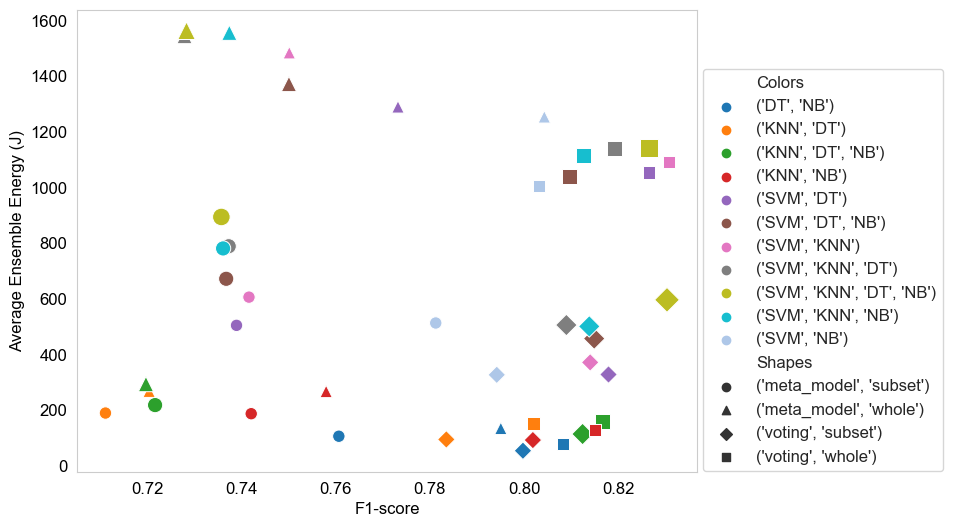} 
    \caption{Ensemble trade-offs: accuracy vs. energy consumption}
    \label{fig:discussion-accuracy-energy}
\end{figure}

We summarize and visualize our results in Fig.~\ref{fig:discussion-accuracy-energy}, which categorizes our ensembles into four groups based on the four combinations of fusion and partition methods.
These groups also have clear indications for accuracy and energy consumption.
The size of the shapes is adjusted based on the size of the ensembles; the larger the ensemble size, the bigger the shape.

\begin{enumerate}
    \item \textbf{Meta-Model, Whole-Dataset (triangle shape)}: Ensembles trained on the whole dataset and fused with a meta-model consume more energy and offer lower accuracy compared to other categories. They are mostly located in the top-left quadrant of Fig.~\ref{fig:discussion-accuracy-energy}.
    \item \textbf{Majority voting, Whole-Dataset (square shape)}: Ensembles trained on the whole dataset and fused using majority voting consume less energy than the triangle shape but more than the circle and diamond shapes. However, they offer the highest accuracy. These ensembles are mostly in the top-right quadrant.
    \item \textbf{Meta-Model, Subset (circle shape)}: Ensembles trained on subsets of the dataset and fused with a meta-model consume less energy than the triangle and square shapes but offer the lowest accuracy of all. They are mostly located in the bottom-left quadrant.
    \item \textbf{Majority voting, Subset (diamond shape)}: Ensembles trained on subsets and fused using majority voting consume the least energy and offer higher accuracy than the triangle and circle shapes but slightly less than the square shape. They are all located in the bottom-right quadrant.
\end{enumerate}

From a Green AI perspective, the ensembles shown as diamond shapes (Majority voting, Subset) are the most preferable.
This category consumed the least energy and offered only a marginally lower accuracy than the (Majority voting, Whole) category, i.e., an average difference of less than 0.02, which is negligible in several cases.
We see that ensembles are clustered based on fusion and partitioning methods rather than ensemble size, indicating that these factors are more influential.
However, within these clusters, we still see the impact of size on energy consumption, i.e., ensembles with more models tend to consume more energy.
While size has a positive effect on the accuracy of ensembles only in categories 2 and 4, it is notable that in both categories, the fusion method is majority voting.
Interestingly, it has a more positive effect on accuracy in category 4 compared to category 2.
In category 4, the average ensemble of size 2 offers 0.802 accuracy, an ensemble of size 3 offers 0.813 accuracy, and an ensemble of size 4 offers 0.830 accuracy.
This shows that an ensemble of size 4 offers a 0.017 increase in accuracy compared to size 3 and a 0.028 increase in accuracy compared to size 2.
Additionally, size 3 offers a 0.011 increase in accuracy compared to an ensemble of size 2.

Some of the ensembles in category 2 with smaller sizes offer almost the same accuracy as the size 4 ensembles in category 4, even the two ensembles (SVM, DT) and (SVM, KNN) in category 2 with smaller sizes that offer equal (SVM, KNN) accuracy or very close (SVM, DT) to the ensemble of size 4 in category 4 consume more energy because the models in category 2 are trained on the whole dataset.
However, there are exceptions, as seen in Fig.~\ref{fig:rq1-energy-vs-size} and Fig.~\ref{fig:rq1-accuracy-vs-size}.
For example, a (KNN, DT, NB) ensemble consumes less energy than some size 2 ensembles, such as (SVM, NB), because the energy consumption depends on the individual models.
SVM consumes more energy than KNN and NB combined, illustrating the importance of model selection.

Regarding accuracy, ensembles of the same size offer different accuracy levels.
Analyzing the correlation between individual accuracy and its effect on the overall accuracy of the model, we observe that there is no direct correlation.
The average F1-score is 0.516 for Naive Bayes, 0.773 for the decision tree, 0.796 for k-nearest neighbors, and 0.800 for the support vector machine.
Naive Bayes and the decision tree offer less accuracy individually compared to the other two models, but the (NB, DT) ensemble surprisingly offers \textit{higher} accuracy than the (SVM, KNN) ensemble.
Deciding which models to include in the ensemble is another important issue that affects both the energy consumption and accuracy of the ensemble.
Different methods are proposed to select the right number of models in an ensemble, which can be roughly divided into clustering-based methods, ranking-based methods, and selection-based methods~\citep{yang2023survey}.

As discussed in RQ2, meta-model fusion consumes more energy compared to majority majority voting, and the energy difference decreases as the number of models increases. On average, meta-model consume 36.16\% more energy than majority voting for size 2 ensembles, 33.18\%  more energy for size 3, and 29.06\% more energy for size 4. As the number of models increases in the ensemble, the percentage difference decreases.
This is because the energy consumed by the fusion process becomes a smaller part of the total energy consumption as the number of models increases.
Conversely, the absolute energy difference increases due to the greater number of features in the meta-model, which consumes more energy to train.
The choice of meta-model impacts the energy and accuracy of ensembles.
We selected K-nearest neighbors (KNN) as it is the most energy-efficient model among the four selected models and offers the second-highest accuracy with the selected datasets.

As KNN is one of the most energy-efficient models among the available traditional ML models for classification tasks~\citep{verdecchia2022data}, we can generalize that the meta-model approach consumes more energy than the majority voting method.
However, the accuracy of a model is entirely dependent on the datasets it uses; there is no inherently good or bad model~\citep{sheth2022comparative}.
Even though meta-model fusion led to accuracy improvements in previous work~\cite{Bonissone2011}, this was not the case in our experiments, most likely because of the small number of models in the ensemble.
This limited the number of features the meta-model could use for predictions, and therefore rendered it ineffective.
Moreover, parameter tuning can also affect the meta-model's efficiency, which was beyond the scope of our study.
Therefore, we can generalize that, with a limited number of models, the meta-model option could be referred to as a never-recommended option because of its higher energy consumption and lower accuracy compared to majority voting.

As seen in Fig.~\ref{fig:discussion-accuracy-energy}, there is an obvious difference between the energy consumption of ensembles trained on the whole dataset vs. a subset.
As training on a subset can have different effects on different models~\citep{verdecchia2022data} and different models have different energy requirements, that is likely of the reasons why we saw different energy savings in different ensembles of the same size.
The method for subset selection is another factor that may also affect the energy consumption and accuracy of the ensemble.
We selected the subsets based on a cross-validation partition of the training data~\citep{zhang2012ensemble,wolpert1992stacked}.
We divided the dataset into mutually exclusive subsets while ensuring that each training instance is used at least once in training an individual model.
There are other methods also used for training the ensemble, such as random sampling with replacement and random sampling without replacement~\citep{sagi2018ensemble}.
Further investigation is needed to compare these different methods.

\textbf{Implications and Recommendations:}
The findings from this study emphasize that it is possible to design ensemble learning components that consume substantially less energy while simultaneously offering comparable or only slightly decreased accuracy.
Increasing the number of models and using whole-dataset training can lead to higher energy costs with only marginal accuracy improvements.
The majority voting fusion method emerged as a particularly effective approach, balancing energy consumption and accuracy.
For practical applications, it is recommended to use majority voting fusion and subset-based training methods to reduce energy consumption without sacrificing accuracy.
Additionally, limiting the ensemble size to what is necessary to achieve acceptable performance and consciously selecting suitable model types can further enhance energy efficiency.
As energy-efficient ML algorithms, we especially recommend experimenting with decision tree, Naive Bayes, and KNN, as ensembles with these algorithms required very little energy in our experiment, while still providing convincing accuracy.
Future work could explore this impact of different types of models within ensembles further, and investigate more sophisticated fusion methods that may offer even better energy efficiency.

\begin{tcolorbox}[colframe=black, colback=white, sharp corners]
\textbf{Summary:} \\
Increasing ensemble size and using whole-dataset training methods significantly increase energy consumption with negligible accuracy improvements. The majority voting fusion method and subset-based training are recommended for their balance of energy consumption and accuracy.
\end{tcolorbox}

\section{Threats to Validity}
A potential threat to \textbf{internal validity}, associated with historical factors, may have emerged in our experiment due to the impact of executing successive iterations on our measurements, such as increasing hardware temperatures.
To mitigate this concern, we introduced a 5-second sleep operation before each experimental iteration, ensuring more uniform hardware conditions for all runs.
Additionally, a warm-up operation was performed to ensure that the initial iteration occurred under conditions similar to subsequent ones, reducing potential measurement influences. 
Regarding the energy measures, the presence of background tasks during the experiment could have acted as confounding factors that affected energy measurements.
To address this, we terminated non-essential processes and restricted access to the infrastructure.
Furthermore, we conducted each experiment iteration 30 times to minimize the impact of any unforeseen background processes.

Concerning \textbf{external validity}, i.e., the generalizability of our findings, we carefully selected tabular classification datasets with different characteristics to have some variety.
We also included four commonly used base classifiers to enhance the study's diversity.
This deliberate selection of datasets and classifiers contributes to a more robust evaluation of the generalizability and effectiveness of the examined methods.
Nonetheless, our results cannot be easily transferred to, e.g., deep learning, regression, or larger number of models in the ensemble.
Additional research is required to address this.

Regarding \textbf{reliability}, we have made a replication package available online to ensure the reproducibility of our study.\footnote{\url{https://zenodo.org/doi/10.5281/zenodo.11664165}}
Additionally, conducting the experiments on different hardware yielded consistent results, reinforcing the reliability of our findings and ensuring the robustness of the outcomes.
The exclusive use of CodeCarbon to estimate energy consumption could threaten the construct validity of our experiment.
To mitigate this risk, we took two steps.
First, we leveraged its open-source nature to examine its implementation and verify its use of RAPL on Linux for gathering energy data.
Second, we conducted the experiment on two different systems to confirm consistent results.

\section{Conclusion}
Our controlled experiment provided a comprehensive analysis of the impact of several design decisions on energy consumption and accuracy in ensemble learning.
Increasing the number of models in an ensemble led to significantly higher energy consumption, with no substantial gains in accuracy.
This highlights the inefficiency of enlarging ensembles indiscriminately.
Comparative analysis of fusion methods revealed that the majority voting approach is notably less energy-hungry than the meta-model, while also providing slightly higher accuracy.
Further, the comparison of training methods showed that while whole-dataset training consumes considerably more energy, it does not offer meaningful improvements in accuracy over subset-based training.
This underscores the potential for significant energy savings through data-centric approaches.
Overall, these findings emphasize the need for careful consideration of energy efficiency in the design and implementation of ensemble learning systems.
From a Green AI perspective, we recommend building ensembles of small size that use majority voting for fusion, subset-based training, and rely on energy-efficient ML algorithms like decision trees, Naive Bayes, or KNN.
However, future research is required to investigate other types of ensembles and forms of machine learning, such as deep learning, to continue exploring innovative ways to optimize energy efficiency of ML-enabled systems.

\section{Acknowledgement}
This work is partially supported by the PON scholarship of Italian government. 

\bibliographystyle{elsarticle-harv}
\bibliography{references}

\end{document}